\documentclass[10pt,twocolumn,letterpaper]{article} 

\usepackage{etex}

\usepackage[iccv_submission]{optional}

\opt{iccv_submission}{
  \usepackage{iccv}
  \usepackage{times}
  \usepackage{epsfig}
  \usepackage{graphicx}
  \usepackage{nicefrac}

  \usepackage{amsmath}
  \usepackage{amssymb}

  \definecolor{ggreen}{HTML}{115740}

  \usepackage[pagebackref=true,breaklinks=true,letterpaper=true,colorlinks=true,bookmarks=false,linkcolor=ggreen,citecolor=ggreen]{hyperref}
  
  \iccvfinalcopy 
  
  \def\iccvPaperID{****} 

  \ificcvfinal\pagestyle{empty}\fi
}

\opt{icml_submission,for_gig}{
  \usepackage{graphicx}
  \usepackage{amsmath}
  \usepackage{amssymb}
  \usepackage[colorlinks=true,linkcolor=blue,citecolor=blue]{hyperref}
  \usepackage[table]{xcolor}
}

\usepackage{microtype}
\usepackage{subfigure}
\usepackage{booktabs} 
\usepackage{enumitem}
\usepackage{amsfonts}
\usepackage{tikz}

\usetikzlibrary{arrows.meta, 
                bending,     
                patterns     
               }

\usetikzlibrary{spy}
\usetikzlibrary{arrows,shapes,automata,backgrounds,petri,positioning}
\usetikzlibrary{decorations.pathmorphing}
\usetikzlibrary{decorations.shapes}
\usetikzlibrary{decorations.text}
\usetikzlibrary{decorations.fractals}
\usetikzlibrary{decorations.footprints}
\usetikzlibrary{shadows}

\newcommand{\fxpsi}{\Phi_{\theta}^{BA}}
\newcommand{\fxvarphi}{\Phi_{\theta}^{AB}}
\newcommand{\fxpsivarepsilon}{\Phi_{\theta \varepsilon}^{BA}}
\newcommand{\fxvarphivarepsilon}{\Phi_{\theta \varepsilon}^{AB}}
\graphicspath{{figs/}}

\newcommand{\mn}[1]{{\color{black}{#1}}}

\opt{icml_submission}{
\usepackage{icml2021}

\usepackage{setspace}
\usepackage[labelfont={bf,small},font={small,stretch=1.1}]{caption}

\usepackage[colorinlistoftodos,prependcaption,textsize=tiny]{todonotes}
}

\opt{icml_accepted}{
  \usepackage[accepted]{icml2021}
}

\opt{for_gig}{
  \usepackage[accepted]{icml2021_unpublished_draft}
  \usepackage[colorinlistoftodos,prependcaption,textsize=tiny,disable]{todonotes}
}

\usepackage[labelfont={bf,small},font={small}]{caption}
\begin{document}

\sloppy

\opt{iccv_submission}{
\title{ICON: Learning Regular Maps Through Inverse Consistency}

\author{Hastings Greer\\
  Department of Computer Science\\
UNC Chapel Hill, USA\\
{\tt\small tgreer@cs.unc.edu}
\and
Roland Kwitt\\
Department of Computer Science\\
University of Salzburg, Austria\\
{\tt\small Roland.Kwitt@sbg.ac.at}
\and
Fran\c{c}ois-Xavier Vialard\\
LIGM, Universit\'e Gustave Eiffel, France\\
{\tt\small francois-xavier.vialard@u-pem.fr}
\and
Marc Niethammer\\
 Department of Computer Science\\
UNC Chapel Hill, USA\\
{\tt\small mn@cs.unc.edu}
}

\maketitle

\thispagestyle{plain}
\pagestyle{plain}

\ificcvfinal\thispagestyle{empty}\fi
}

\opt{for_gig,icml_submission}{

\icmltitlerunning{Learning Regular Maps Through Inverse Consistency}
  
\twocolumn[
\icmltitle{Learning Regular Maps Through Inverse Consistency}



\icmlsetsymbol{equal}{*}

\begin{icmlauthorlist}
\icmlauthor{Hastings Greer}{unc}
\icmlauthor{Roland Kwitt}{salzburg}
\icmlauthor{Fran\c{c}ois Xavier-Vialard}{upem}
\icmlauthor{Marc Niethammer}{unc}
\end{icmlauthorlist}

\icmlaffiliation{unc}{Department of Computer Science, University of North Carolina at Chapel Hill, USA}
\icmlaffiliation{salzburg}{Department of Computer Science, University of Salzburg, Austria}
\icmlaffiliation{upem}{LIGM, Universit\'e Gustave Eiffel, France}

\icmlcorrespondingauthor{Hastings Greer}{tgreer@cs.unc.edu}

\icmlkeywords{Deep Learning, Image Registration, Inverse Consistency}

\vskip 0.3in
]



\printAffiliationsAndNotice{\icmlEqualContribution} 
}

\begin{abstract}
  Learning maps between data samples is fundamental. Applications range from representation learning, image translation and generative modeling, to the estimation of spatial deformations. Such maps relate feature vectors, or map between feature spaces. Well-behaved maps should be regular, which can be imposed explicitly or may emanate from the data itself. We explore what induces regularity for spatial transformations, e.g., when computing image registrations. Classical optimization-based models compute maps between pairs of samples and rely on an appropriate regularizer for well-posedness. Recent deep learning approaches have attempted to avoid using such regularizers altogether by relying on the sample population instead. We explore if it is possible to obtain spatial regularity using an inverse consistency loss only and elucidate what explains map regularity in such a context. We find that deep networks combined with an inverse consistency loss and randomized off-grid interpolation yield well behaved, approximately diffeomorphic, spatial transformations. Despite the simplicity of this approach, our experiments present compelling evidence, on both synthetic and real data, that regular maps can be obtained without carefully tuned explicit regularizers, \mn{while achieving competitive registration performance.}
\end{abstract}

\section{Motivation}
\label{section:motivation}

Learning maps between feature vectors or spaces is an important task. Feature vector maps are used to improve representation learning~\cite{chen2020exploring}, or to learn correspondences in natural language processing~\cite{blitzer2006domain}. Maps between spaces are important for generative models when using normalizing flows~\cite{kobyzev2020normalizing} (to map between a simple and a complex probability distribution), or to determine spatial correspondences between images, \eg, for optical flow~\cite{horn1981determining} to determine motion from videos~\cite{fortun2015optical}, depth estimation from stereo images~\cite{laga2020survey}, or medical image registration~\cite{sotiras2013deformable,tustison2019learning}.

\vskip0.5ex
Regular maps are typically desired; \eg, diffeomorphic maps for normalizing flows to properly map densities, or for medical image registration to map to an atlas space~\cite{joshi2004unbiased}. Estimating such maps requires an appropriate choice of transformation model. This entails picking a parameterization, which can be simple and depend on few parameters (\eg, an affine transformation), or which can have millions of parameters for 3D nonparametric approaches~\cite{holden2007review}. Regularity is achieved by 1) picking a simple transformation model with limited degrees of freedom, 2) regularization of the transformation parameters, 3) or implicitly through the data itself. Our goal is to demonstrate and understand how spatial regularity of a transformation can be achieved by encouraging \emph{inverse consistency} of a map. Our motivating example is image registration/optical flow, but our results are applicable to other tasks where spatial transformations are sought. 


\vskip0.5ex
Registration problems have traditionally been solved by numerical optimization~\cite{modersitzki2004numerical} of a loss function balancing an image similarity measure and a regularizer. Here, the predominant paradigm is \emph{pair-wise} image registration\footnote{A notable exception is congealing~\cite{zollei2005efficient}.} where many maps may yield good image similarities between a transformed moving and a fixed image; the regularizer is required for well-posedness to single out the most desirable map. Many different regularizers have been proposed~\cite{holden2007review,modersitzki2004numerical,risser2011simultaneous} and many have multiple hyperparameters, making regularizer choice and tuning difficult in practice. Deep learning approaches to image registration and optical flow have moved to learning maps from \emph{many image pairs}, which raises the question if explicit spatial regularization is still required, or if it will emanate as a consequence of learning over many image pairs. For optical flow, encouraging results have been obtained without using a spatial regularizer~\cite{dosovitskiy2015flownet,ranjan2017optical}, though more recent work has advocated for spatial regularization to avoid ``vague flow boundaries and undesired artifacts''~\cite{hui2018liteflownet,hur2019iterative}. Interestingly, for medical image registration, where map regularity is often very important, almost all the existing work uses regularizers as initially proposed for pairwise image registration~\cite{shen19,yang2017quicksilver,balakrishnan2019voxelmorph} with the notable exception of~\cite{bhalodia19} where the deformation space is guided by an autoencoder instead.

\vskip0.5ex
Limited work explores if regularization for deep registration networks can be avoided entirely, or if weaker forms of regularizations might be sufficient. To help investigate this question, we work with binary shapes (where regularization is particularly important due to the aperture effect~\cite{horn1986robot}) and real images. We show that regularization is necessary, but that carefully encouraging \emph{inverse consistency} of a map suffices to obtain approximate diffeomorphisms. The result is a simple, yet effective, nonparametric approach to obtain well-behaved maps, which only requires limited tuning. In particular, the in practice often highly challenging process of selecting a spatial regularizer is eliminated.


\vskip0.5ex
\noindent {\bf Our contributions} are as follows: (1) We show that \emph{approximate} inverse consistency, combined with off-grid interpolation, results in approximate diffeomorphisms, when using a deep registration model trained on large datasets. Foregoing regularization is insufficient; (2) Bottleneck layers are not required and many network architectures are suitable; (3) Affine preregistration is \mn{not required}; (4) We propose randomly sampled evaluations to avoid transformation flips in texture-less areas and an inverse consistency loss with beneficial boundary effects; (5) We present good results of our approach on synthetic data, MNIST, and a 3D magnetic resonance knee dataset \mn{of the Osteoarthritis Initiative (OAI).}



\section{Background and Analysis}
\label{section:background}

Image registration is typically based on solving  optimization problems of the form
\begin{equation}
  \theta^* = \underset{\theta}{\text{argmin}}~\mathcal{L}_{\text{sim}}(I^A\circ \Phi^{-1}_\theta,I^B) + \lambda\mathcal{L}_{\text{reg}}(\theta)\enspace,\label{eq:basic_reg}
\end{equation}
where $I^A$ and $I^B$ are moving and fixed images, $\mathcal{L}_{\text{sim}}(\cdot,\cdot)$ is the similarity measure, $\mathcal{L}_{\text{reg}}(\cdot)$ is a regularizer
, $\theta$ are the transformation parameters, $\Phi_\theta$ is the transformation map, and $\lambda\geq 0$. We consider images as functions from $\mathbb{R}^N$ to $\mathbb{R}$ and maps as functions from $\mathbb{R}^N$ to $\mathbb{R}^N$. We write $\|f\|_p$ for the $L^p$ norm on a scalar or vector-valued function $f$.

\vskip0.5ex
Maps, $\Phi_\theta$, can be parameterized using few parameters (\eg, affine, B-spline~\cite{holden2007review}) or nonparametrically with continuous vector fields~\cite{modersitzki2004numerical}. In the nonparametric case, parameterizations are infinite-dimensional (as one deals with function spaces) and represent displacement, velocity, or momentum fields~\cite{balakrishnan2019voxelmorph,shen19,yang2017quicksilver,modersitzki2004numerical}. Solutions to Eq.~\eqref{eq:basic_reg} are classically obtained via numerical optimization~\cite{modersitzki2004numerical}. Recent deep registration networks are conceptually similar, but \emph{predict} $\tilde{\theta}^*$, \ie, an estimate of the true minimizer $\theta^*$. 

\vskip0.5ex
There are three interesting observations: \emph{First}, for transformation models with few parameters (\eg, affine), regularization is often not used (\ie, $\lambda=0$). \emph{Second}, while deep learning (DL) models minimize losses similar to Eq.~\eqref{eq:basic_reg}, the parameterization is different: it is over network \emph{weights}, resulting in a predicted $\theta^*$ instead of optimizing over $\theta$ directly. \emph{Third}, DL models are trained over \emph{large collections of image pairs} instead of a single $(I^A,I^B)$ pair. This raises the following questions: \textbf{Q1}) Is explicit spatial regularization necessary, or can we avoid it for nonparametric registration models? \textbf{Q2}) Is using a \emph{single} neural network parameterization to predict \emph{all} $\theta^*$ beneficial? For instance, will it result in simple solutions as witnessed for deep networks on other tasks~\cite{shah2020pitfalls} or capture meaningful deformation spaces as observed in~\cite{yang2017quicksilver}? \textbf{Q3}) 
Does a deep network parameterization itself result in regular solutions, even if only applied to a single image pair, as such effects have, \eg, been observed for structural optimization~\cite{hoyer2019neural}?

\vskip0.5ex
Regularization typically encourages spatial smoothness by penalizing derivatives (or smoothing in dual space). Commonly, one uses a Sobolev norm or total variation. Ideally, one would like a regularizer adapted to deformations one expects to see (as it encodes a prior on expected deformations \eg, as in~\cite{niethammer2019metric}). In consequence, picking and tuning a regularizer is cumbersome and often involves many hyperparameters. While avoiding explicit regularization has been explored for deep registration / optical flow networks~\cite{dosovitskiy2015flownet,ranjan2017optical},  there is evidence that regularization is beneficial~\cite{hui2018liteflownet}. 

\vskip0.5ex
\emph{Our key idea is to avoid complex spatial regularization and to instead obtain approximate diffeomorphisms by encouraging inverse consistent maps \mn{via regularization}.}


\subsection{Weakly-regularized registration}
\label{subsection:regularization_thoughts_and_sorting}

Assume we eliminate regularization ($\lambda=0$) and use the $p$-th power of the $L^p$ norm of the difference between the warped image, $I^A\circ\Phi_\theta^{-1}$, and the fixed image, $I^B$, as similarity measure. Then, our optimization problem becomes
\begin{equation}
  \theta^* = \underset{\theta}{\arg \min}~\int (I^A(\Phi_\theta^{-1}(x))-I^B(x))^p~\mathrm{d}x\,,~p\geq 1, \label{eq:only_similarity}
\end{equation}
\ie, the image intensities of $I^A$ should be close to the image intensities of $I^B$ \emph{after} deformation. Without regularization, we are entirely free to choose $\Phi_\theta$. Highly irregular minimizers of Eq.~\eqref{eq:only_similarity} may result as each intensity value $I^A$ is simply matched to the closest intensity value of $I^B$ regardless of location. For instance, for a constant $I^B(x)=c$ and a moving image $I^A(y)$ with a unique location $y_c$, where $I^A(y_c)=c$, the optimal map is $\Phi_\theta^{-1}(x) = y_c$, which  is not invertible: only \emph{one point} of $I^A$ will be mapped to the \emph{entire} domain of $I^B$. Clearly, more spatial regularity is desirable. Importantly, irregular deformations are common optimizers of Eq.~\eqref{eq:only_similarity}. 

\begin{figure}
  \centering
  \includegraphics[width=0.98\columnwidth]{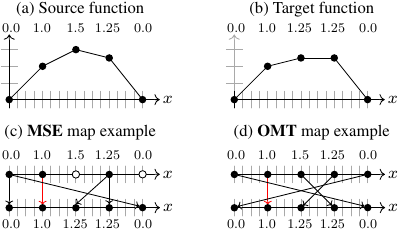}
  \caption{Source and target functions for a 1D registration example. Panels (c) and (d) show two possible solutions for mean square error (MSE) and OMT, respectively. In both cases,  solutions may not be unique. However, for OMT, matching solutions will be one-to-one, \ie, invertible. OMT imposes a stronger constraint than MSE on the obtainable maps, but irregular maps are still permissible.
  }
  \label{fig:matching_examples}
  \vspace{-0.2cm}
\end{figure}

\vskip1ex
Optimal mass transport (OMT) is widely used in machine learning and in imaging. Such models are of interest to us as they can be inverse consistent. An OMT variant of the discrete reformulation of Eq.~\eqref{eq:only_similarity} is
\begin{equation}
    \theta^* = \underset{\theta}{\arg\min}~\mathrm{d}x \sum_{i=1}^S (I^A(\Phi_\theta^{-1}(x_i))-I^B(x_i))^p\,, p\geq 1\label{eq:only_similarity_discrete}
\end{equation}
for $p\geq 1$, where $i$ indexes the $S$ grid points $x_i$, $\Phi_\theta^{-1}(x_i)$ is restricted to map to the grid \mn{points} $y_i$ of $I^A$, and $\mathrm{d}x$ is the discrete area element. Instead of considering all possible maps, we attach a unit mass to each \emph{intensity} value of $I^A$ and $I^B$ and ask for minimizers of Eq.~\eqref{eq:only_similarity_discrete} which transform the intensity distribution of $I^A$ to the intensity distribution of $I^B$ via \emph{permutations} of the values only. As we only allow permutations, the optimal map will be \emph{invertible} by construction. This problem is equivalent to optimal mass transport for one-dimensional empirical measures~\cite{peyre2019computational}. One obtains the optimal value by ordering all intensity values of $I^A$ ($I^A_1\leq \cdots\leq I^A_S$) and $I^B$ ($I^B_1\leq \cdots\leq I^B_S$). The minimum is the $p$-th power of the  $p$-Wasserstein distance ($p\geq 1$) 
$\mathcal{W}_p^p = \sum_i |I^A_i-I^B_i|^p$.
In consequence, minimizers for Eq.~\eqref{eq:only_similarity} are related to sorting, but do not consider spatial regularity. Note that solutions might not be unique when intensity values in $I^A$ or $I^B$ are repeated. 
Solutions via sorting were empirically explored for registration in~\cite{Rholfing12} to illustrate that they, in general, do not result in spatially meaningful registrations.
At this point, our idea of using inverse consistency (\ie, invertible maps)
as the only regularizer appears questionable, given that OMT often provides an inverse consistent model (when a matching, \ie, a Monge solution, is optimal), while resulting in irregular maps (Fig.~\ref{fig:matching_examples}).

\vskip0.5ex
\emph{\mn{Yet, we will show that a registration network, combined with 
  an inverse consistency loss, encourages map regularity.}}


\subsection{Avoiding undesirable solutions} 
\label{subsec:deep_network_intuition}

\noindent
\textbf{Simplicity}.
The highly irregular maps in Fig.~\ref{fig:matching_examples} occur for \emph{pair-wise} image registration. Instead, we are concerned with training a network over an \emph{entire image population}. Were one to find a global inverse consistent minimizer, a network would need to implicitly approximate the sorting-based OMT solution. As sorting is a continuous piece-wise linear function~\cite{blondel2020fast}, it can, in principle, be approximated according to the universal approximation theorem~\cite{leshno1993multilayer}. However, this is a limit argument. Practical neural networks for sorting are either \emph{approximate}~\cite{liu2011learning,engilberge2019sodeep} or very large (\eg, $O(S^2)$ neurons for $S$ values~\cite{chen1990neural}). Note that deep networks often tend to simple solutions~\cite{shah2020pitfalls} and that we do not even want to sort \emph{all} values for registration. Instead, we are interested in more \emph{local} permutations, rather than the global OMT permutations, which is what we will obtain for neural network solutions with inverse consistency.

\vskip0.5ex
\noindent
\textbf{Invertibility}.
Requiring map invertibility implies searching for a matching (a Monge formulation in OMT) which is an optimal permutation, but which may not be continuous\footnote{It would be interesting to study how well a network approximates an OMT solution and if it naturally regularizes it.}.
Instead, our goal is a \emph{continuous and invertible} map. 
We therefore want to penalize deviations from
\begin{equation}
    \Phi_{\theta}^{AB} \circ \Phi_{\theta}^{BA} = \operatorname{Id}\,,\label{eq:inverse_consistency}
\end{equation}
where $\Phi_{\theta}^{AB}$ denotes a predicted map (by a network with weights $\theta$) to register image $I^A$ to $I^B$; $\Phi_{\theta}^{BA}$ is the network output with reversed inputs and $\operatorname{Id}$ denotes the identity map.

\vskip0.5ex
Inverse consistency of maps has been explored to obtain symmetric maps for pair-wise registration~\cite{hart2009optimal,christensen2001consistent} and for registration networks~\cite{zhang18,shen19}. Related losses have been proposed on images (instead of maps) for registration~\cite{boah19,boah20} and for image translation~\cite{zhang2020cross}. However, none of these approaches study inverse consistency for regularization. Likely, because it has so far been believed that additional spatial regularization is required for nonparametric registration.

\subsection{Approximate inverse consistency} 
\label{subsec:h1_by_approximate_inverse_consistency}

As we will show next, \emph{approximate inverse consistency} by itself yields regularizing effects in the context of pairwise image registration.

\vskip0.5ex
Denote by $\fxvarphi(x)$ and $\fxpsi(x)$ the output maps of a network for images 
$(I^A,I^B)$ and $(I^B,I^A)$, respectively.
As inverse consistency by itself does not prevent discontinuous solutions, we propose to use \emph{approximate} inverse consistency to favor $C^0$ solutions. 
We add two vector-valued independent spatial white noises $n_1(x),n_2(x)\in\mathbb{R}^N$
($x \in [0,1]^N$ with $N$=2 or $N$=3 the image dim.) of variance $1$ for each space location and dimension to the two output maps and define 
\begin{align*}
  \fxvarphivarepsilon(x) & = \fxvarphi(x) + \varepsilon n_1(\fxvarphi(x))\enspace, \\ 
  \fxpsivarepsilon(x)    & = \fxpsi(x) + \varepsilon n_2(\fxpsi(x))\enspace,
\end{align*}
with $\varepsilon>0$. We then consider the loss $\mathcal{L} = \lambda\mathcal{L}_{\text{inv}} + \mathcal{L}_{\text{sim}}$, with inverse consistency component ($\mathcal{L}_{\text{inv}}$)
\begin{equation}
\begin{split}
  \mathcal{L}_{\text{inv}} & = 
  \left\| \fxvarphivarepsilon \circ \fxpsivarepsilon - \operatorname{Id} \right\|^2_2 + 
  \left\| \fxpsivarepsilon \circ \fxvarphivarepsilon - \operatorname{Id} \right\|^2_2
\end{split}
\label{EqLossSymmetric:partInv}
\end{equation}
and similarity component ($\mathcal{L}_{\text{sim}}$)

\begin{equation}
  \mathcal{L}_{\text{sim}}  =  
  \left\| I^A \circ \fxvarphi - I^B \right\|^2_2 + 
  \left\| I^B \circ \fxpsi  - I^A \right\|^2_2\enspace.
  \label{EqLossSymmetric:partSim}
\end{equation}
Importantly, note that there are \emph{multiple} maps that can lead to the same $I^A \circ \fxvarphi$ and $I^B \circ \fxpsi$. Therefore, among all these maps,  minimizing the loss $\mathcal{L}$ drives the maps towards those that minimize the two terms in Eq.~\eqref{EqLossSymmetric:partInv}.
%
%
%

\vskip1ex
\noindent
\textbf{Assumption}.
\emph{Both terms in Eq.~\eqref{EqLossSymmetric:partInv} can be driven to a small value (of the order of the noise), by minimization.}
\par
%

\vskip1ex
We first Taylor-expand one of the two terms in Eq.~\eqref{EqLossSymmetric:partInv} (the other follows similarly), yielding
\begin{equation}
  \begin{split}
  \left\| \fxvarphivarepsilon \circ \fxpsivarepsilon - \operatorname{Id} \right\|^2_2  \approx
  & \left\| \fxvarphi \circ \fxpsi~+ \right. \\
  & ~~~~\varepsilon n_1(\fxvarphi \circ \fxpsi)~+ \\
  & ~~~\left. \mathrm{d}\fxvarphivarepsilon(\varepsilon n_2(\fxpsi)) - \operatorname{Id}\right\|^2_2\enspace.\nonumber
  \end{split}
\end{equation}
Defining the right-hand side as $A$, developing the squares and taking expectation, we obtain 
\begin{equation}
  \begin{split}
   \mathbb{E}[A] = & \left\| \fxvarphi \circ \fxpsi - \operatorname{Id} \right\|^2_2  \\
   & + \varepsilon^2 \mathbb{E}\left[\left\|n_1\circ(\fxvarphivarepsilon \circ \fxpsivarepsilon)\right\|^2_2\right] \\ 
   & + \varepsilon^2\mathbb{E}\left[\left\|\mathrm{d}\fxvarphivarepsilon( n_2) \circ \fxpsi \right\|^2_2\right]\enspace,
  \end{split}
  \label{eqn:expectation_of_A}
\end{equation}
since, by independence, all the cross-terms vanish (the noise terms have $0$ mean value).
The second term is constant, \ie,
\begin{alignat}{1}
  &\mathbb{E}\left[\left\|n_1\circ(\fxvarphivarepsilon \circ \fxpsivarepsilon)\right\|^2_2\right] =
  \\
  &\int \mathbb{E}\left[\|n_1\|^2_2(y)\right] \operatorname{Jac}((\fxpsivarepsilon)^{-1} \circ (\fxvarphivarepsilon)^{-1})~\mathrm{d}y= \text{const.} \,, \notag
\end{alignat}
where we performed a change of variables and denoted the determinant of the Jacobian matrix as $\operatorname{Jac}$. The last equality follows from the fact that the variance of the noise term is spatially constant and equal to $1$. 
By similar arguments, the last expectation term in Eq.~\eqref{eqn:expectation_of_A} can be rewritten as
\begin{multline}\label{EqWhiteNoise}
  \mathbb{E}\left[\left\|\mathrm{d}\fxvarphivarepsilon( n_2) \circ \fxpsi \right\|^2_2\right] = \\ 
   \int \operatorname{Tr}(\mathrm{d}(\fxvarphivarepsilon)^{\top} \mathrm{d}\fxvarphivarepsilon) \operatorname{Jac}((\fxpsi)^{-1})~\mathrm{d}y\,,
\end{multline}
where $\operatorname{Tr}$ denotes the trace operator. As detailed in the suppl. material, the identity of Eq.~\eqref{EqWhiteNoise} relies on a change of variable and on the property of the white noise, $n_2$, which satisfies null correlation in space and dimension $\mathbb{E}[n_2(x) n_2(x')^\top] = \operatorname{Id}_{\mathbb{R}^N}$ if $x=x'$ and $0$ otherwise.

\vskip0.5ex
\noindent
\textbf{Approximation \& $H^1$ regularization}.
We now want to connect the approximate inverse consistency loss of Eq.~\eqref{EqLossSymmetric:partInv} with $H^1$ norm type regularization. 
Our assumption implies that $\fxvarphi \circ \fxpsi,\fxpsi \circ \fxvarphi$ are close to identity, therefore one has 
$\operatorname{Jac}((\fxpsi)^{-1}) \approx \operatorname{Jac}(\fxvarphi)$. Assuming this approximation holds
, we use it in Eq.~\eqref{EqWhiteNoise}, together with the fact that, $\fxvarphivarepsilon \approx \fxvarphi + O(\varepsilon)$ to get at order $\varepsilon^2$ (see suppl. material for details) to approximate $\mathcal{L}_{\text{inv}}$, \ie,
\begin{equation}
  \begin{split}
  \mathcal{L}_{\text{inv}} & \approx \left\| \fxvarphi \circ \fxpsi - \operatorname{Id}\right\|^2_2  + \left\| \fxpsi \circ \fxvarphi - \operatorname{Id}\right\|^2_2 \\
  + \varepsilon^2 &\left\| d \fxvarphi \sqrt{\operatorname{Jac}(\fxvarphi)} \right\|^2_2 
  + \varepsilon^2 \left\| d \fxpsi \sqrt{\operatorname{Jac}(\fxpsi)} \right\|^2_2 \, 
  \end{split}
\label{EqH1regularization}
\end{equation}
We see that approximate inverse consistency leads to an $L^2$ penalty of the gradient, weighted by the Jacobian of the map. This is a type of Sobolev ($H^1$ more precisely) regularization sometimes used in image registration. In particular, the $H^1$ term is likely to control the compression and expansion magnitude of the maps, at least on average, on the domain.
\mn{\emph{Hence, approximate inverse consistency leads to an implicit $H^1$ regularization, formulated directly on the map.}}

\vskip0.5ex
\noindent
\textbf{Inverse consistency with no noise and the implicit regularization of inverse consistency}.
Turning the noise level to zero also leads to regular displacement fields in our experiments when predicting maps with a neural network. In this case, we observe that inverse consistency is only approximately achieved. Therefore, one can postulate that the error made in computing the inverse entails the $H^1$ regularization as previously shown. 
The possible caveat of this hypothesis is that the inverse consistency error might not be independent of the displacement fields, which was assumed in proving the emerging $H^1$ regularization.
Last, even when the network should have the capacity to exactly satisfy inverse consistency for all data, we conjecture that the implicit bias due to the optimization will favor more regular outputs.

\vskip0.5ex
\emph{A fully rigorous theoretical understanding of the regularization effect due to the data population and its link with inverse consistency 
is important, but beyond our scope here.}

\section{Approximately diffeomorphic registration}

We base our registration approach on training a neural network $F_\theta^{AB}$ which, given input images $I^A$ and $I^B$, outputs a grid of \emph{displacement} vectors, $D_\theta^{AB}$, in the space of image $I^B$, assuming normalized image coordinates covering $[0,1]^N$. We obtain \emph{continuous} maps by interpolation, \ie, 
\begin{equation}
  \Phi_\theta^{AB} = D_{\theta}^{AB} + \operatorname{Id}, \quad D_{\theta}^{AB} = \operatorname{interp}(F_{\theta}^{AB})
  \label{eq:map_interpolation}
\end{equation}
where $I^A\circ\Phi_\theta^{AB} \approx I^B$. Under the assumption of linear interpolation (bilinear in 2D and trilinear in 3D), $\Phi_\theta^{AB}$ is continuous and differentiable except on a measure zero set. Building on the considerations of Sec.~\ref{section:background} we seek to minimize 
\begin{equation}
  \mathcal{L}(\theta) = \mathbb{E}_{p(I^A,I^B)}\left[\mathcal{L}_{\text{sim}}^{AB} + \lambda \mathcal{L}_{\text{inv}}^{AB}\right],\label{eq:overall_loss}
\end{equation}
where $\lambda\geq 0$ and $p(I^A,I^B)$ denotes the distribution over all possible image pairs. The similarity and invertibility losses depend on the neural network parameters, $\theta$, and are 
\begin{alignat}{1}
  \mathcal{L}_{\text{sim}}^{AB} &= \mathcal{L}_{\text{sim}}(I^A \circ \Phi_\theta^{AB}, I^B) + \mathcal{L}_{\text{sim}}(I^B \circ \Phi_\theta^{BA}, I^A) \notag \\
  \mathcal{L}_{\text{inv}}^{AB} &= \mathcal{L}_{\text{inv}}(\Phi_\theta^{AB},\Phi_\theta^{BA}) + \mathcal{L}_{\text{inv}}(\Phi_\theta^{BA},\Phi_\theta^{AB})
\label{eqn:similarity_and_consistency_loss}
\end{alignat}
with
\begin{equation}
    \mathcal{L}_{\text{sim}}(I,J) = \|I-J\|_2^2\,,
    \mathcal{L}_{\text{inv}}(\phi,\psi) = \|\phi \circ \psi - \operatorname{Id}\|_2^2\,.
\end{equation}
For simplicity, we use the squared $L^2$ norm as similarity measure. Other measures, \eg, normalized cross correlation (NCC) or mutual information (MI), can also be used. When $\mathcal{L}_{\text{inv}}^{AB}$ goes to zero, $\Phi_\theta^{AB}$ will be approx. invertible and continuous due to Eq.~\eqref{eq:map_interpolation}. Hence, we obtain approximate $C^0$ diffeomorphisms without  differential equation integration,  hyperparameter tuning, or transform restrictions. Our loss in Eq.~\eqref{eq:overall_loss} is symmetric in the image pairs due to the symmetric similarity and invertibility losses in Eq.~\eqref{eqn:similarity_and_consistency_loss}.



%

\vskip0.5ex
\noindent
\textbf{Displacement-based inverse consistency loss}.
A general map $\Phi_\theta^{AB}$ may map points in $[0, 1]^N$ to points outside $[0, 1]^N$. Extrapolating maps across the boundary is cumbersome. Hence, we only interpolate displacement fields as in Eq.~\eqref{eq:map_interpolation}. We rewrite the inverse consistency loss as
\begin{alignat}{1}
    \mathcal{L}_{\text{inv}}(\Phi_\theta^{AB},\Phi_\theta^{BA}) & = \left\|(D_\theta^{AB} + \operatorname{Id}) \circ (D_\theta^{BA} + \operatorname{Id}) - \operatorname{Id}\right\|^2_2 \notag \\
    & = \left\|(D_\theta^{AB}) \circ \Phi_\theta^{BA}  + D_\theta^{BA} \right\|_2^2
\end{alignat}
and use it for implementation, as it is easier to evaluate.



\begin{figure}
  \centering
  \begin{tabular}{cc}
    \resizebox{0.45\columnwidth}{!}
    {
      \begin{tikzpicture}[spy using outlines={circle,yellow,magnification=4,size=2.0cm, connect spies}]
        \node {\pgfimage[width=0.28\textwidth]{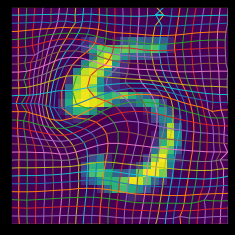}};
        \spy on (0.85,2.1) in node [left] at (3.5,0.75);
      \end{tikzpicture}
    }
    &
    \resizebox{0.45\columnwidth}{!}
    {
      \begin{tikzpicture}[spy using outlines={circle,yellow,magnification=4,size=2.0cm, connect spies}]
        \node {\pgfimage[width=0.28\textwidth]{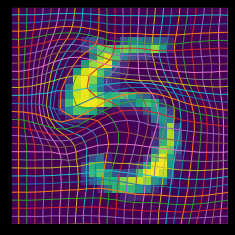}};
        \spy on (0.85,2.1) in node [left] at (3.5,0.75);
      \end{tikzpicture}
    }
    \vspace{-0.1cm}
  \end{tabular}

\caption{The left output is generated by a network trained with inverse consistency, evaluated on a grid instead of randomly. As a result, the loss cannot detect that maps generated by this network flip the pair of pixels in the upper right corner, as that error is not represented in the composed map. The right output is obtained from a network trained with random evaluation off of lattice points.} \label{fig:flips}
\end{figure}


\begin{figure}
  \centering
  \includegraphics[width=\columnwidth]{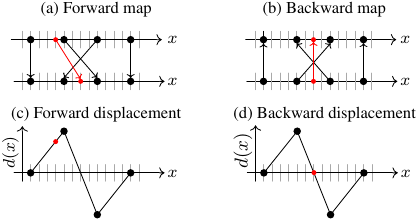}
  \caption{In this example, grid points (solid black discs) map to each other inverse consistently. The forward map (a) is inverted by the backward map (b). However, folding of the space occurs as the middle two points swap positions. Off-grid points map under linear interpolation according to (c/d). We see that the interpolated displacements for the small solid red disc (\textcolor{red}{$\bullet$}) do not result in an invertible map. Hence, this mismatch would be penalized by the inverse consistency loss, but only when evaluated off-grid.}
  \label{fig:off_grid_resampling}
  \vspace{-0.2cm}
\end{figure}

\vskip0.5ex
\noindent
\textbf{Random evaluation of inverse consistency loss}.
$\mathcal{L}_{\text{inv}}^{AB}$ can be evaluated by approximating the $L^2$ norm, assuming constant values over the grid cells. In many cases, this is sufficient. However, as Fig.~\ref{fig:flips} illustrates, swapped locations may occur in uniform regions  where a registration network only sees uniform background. This swap, composed with itself, is the identity as long as it is only evaluated at the center of pixels/voxels. Hence, the map appears invertible to the loss. However, outside the centers of pixels/voxels, the map is not inverse consistent when combined with linear interpolation. To avoid such pathological cases, we approximate the $L^2$ norm by random sampling. This forces interpolation and therefore results in non-zero loss values for swaps. Fig.~\ref{fig:off_grid_resampling} shows why off-grid sampling combined with inverse consistency is a stronger condition than only considering deformations at grid points. In practice, we evaluate the loss

\begin{align}
    & \mathcal{L}_{\text{inv}}(\Phi_\theta^{AB},\Phi_\theta^{BA})\\
    & ~~= \left\| (D_\theta^{AB}) \circ \Phi_\theta^{BA}  + D_\theta^{BA} \right\|_2^2 \nonumber \\
    & ~~= \mathbb{E}_{x \sim \mathcal{U}(0,1)^N} \left[(D_\theta^{AB}) \circ \Phi_\theta^{BA}  + D_\theta^{BA}\right]^2(x) \nonumber \\
    & ~~\approx \nicefrac{1}{N_p}\sum\nolimits_{i} \left([(D_\theta^{AB}) \circ (D_\theta^{BA} + \operatorname{Id})  + D_\theta^{BA}] (x_i + \epsilon_i) \right)^2  \nonumber \\
    & ~~= \nicefrac{1}{N_p}\sum\nolimits_{i} \left([D_\theta^{AB} \circ (D_\theta^{BA} \circ (x_i + \epsilon_i) + x_i + \epsilon_i) \right.  \nonumber \\
    & ~~~~~~~~~~~~~~~~~~~~~~~~+ \left. D_\theta^{BA} \circ (x_i + \epsilon_i)]\right)^2 \nonumber
\end{align}
where $N_p$ is the number of pixels/voxels, $\mathcal{U}(0,1)^N$ denotes the uniform distribution over $[0,1]^N$, $x_i$ denotes the grid center coordinates and $\epsilon_i$ is a random sample drawn from a multivariate Gaussian with standard deviation set to the size of a pixel/voxel in the respective spatial directions.

\section{Experiments}
\label{section:experiments}

Our experiments address several aspects:  First, we compare our approach to \emph{directly} optimizing the maps $\Phi^{AB}$ and $\Phi^{BA}$ on a 2D toy dataset of $128 \times 128$ images. Second, on a 2D toy dataset of $28 \times 28$ images, we assess the impact of architectural and hyperparameter choices. Finally, we assess registration performance on real 3D magnetic resonance images (MRI) of the knee. 

\subsection{Datasets}
\label{subsection:datasets}

\noindent
\textbf{MNIST}. We use the standard MNIST dataset with images of size $28 \times 28$, restricted to the number ``5'' to make sure we have semantically matching images. For training/testing, we rely on the standard partitioning of the dataset.


\vskip0.5ex
\noindent
\textbf{Triangles \& Circles.} We created 2D triangles and circles ($128 \times 128$) with radii and centers varying uniformly in $[.2,.4]$ and $[.4,.7]$, respectively. 
Pixels are set to 1 inside a shape and smoothly decay to -1 on the outside. We train using 6,000 images and test on 6,000 separate images\footnote{Code to generate images and replicate these experiments is available at \url{https://github.com/uncbiag/ICON}}.


\vskip0.5ex
\noindent
\textbf{OAI knee dataset}. These are 3D MR images from the Osteoarthritis Initiative (OAI). Images are downsampled to size $192 \times 192 \times 80$,
normalized such that the 1th percentile is set to 0, the 99th percentile is to 1, and all values are clamped to be in $[0,1]$. As a preprocessing step, images of left knees are mirrored along the left-right axis. The dataset contains ~\mn{2,532} training images and 301 test pairs. 

\subsection{Architectures}
\label{subsection:architectures}

We experiment with four neural network architectures. All networks output displacement fields, $D_\theta^{AB}$. We briefly outline the differences below, but refer to the suppl. material for details. The first network is an \textbf{MLP} with 2 hidden layers and ReLU activations. The output layer is reshaped into size $2 \times W \times H$. 
Second, we use a convolutional encoder-decoder network (\textbf{Enc-Dec}) with 5 layers each, reminiscent of a U-Net \emph{without} skip connections. 
Our third network uses 6 convolutional layers without up- or down-sampling. The input to each layer is the concatenation of the outputs of all previous layers (\textbf{ConvOnly}). 
Finally, we use a \textbf{U-Net} with skip and residual connections. The latter is similar to \textbf{Enc-Dec}, but uses LeakyReLU activations and batch normalization.
In all architectures, the final layer weights are initialized to 0, so that optimization starts at a network outputting a zero displacement field.

\subsection{Regularization by approx. inverse consistency}
\label{subsec:imperfect_inverse_consistency}

\begin{figure}
   \resizebox{\columnwidth}{!}{
     \centering
     \includegraphics[width=1.0\columnwidth]{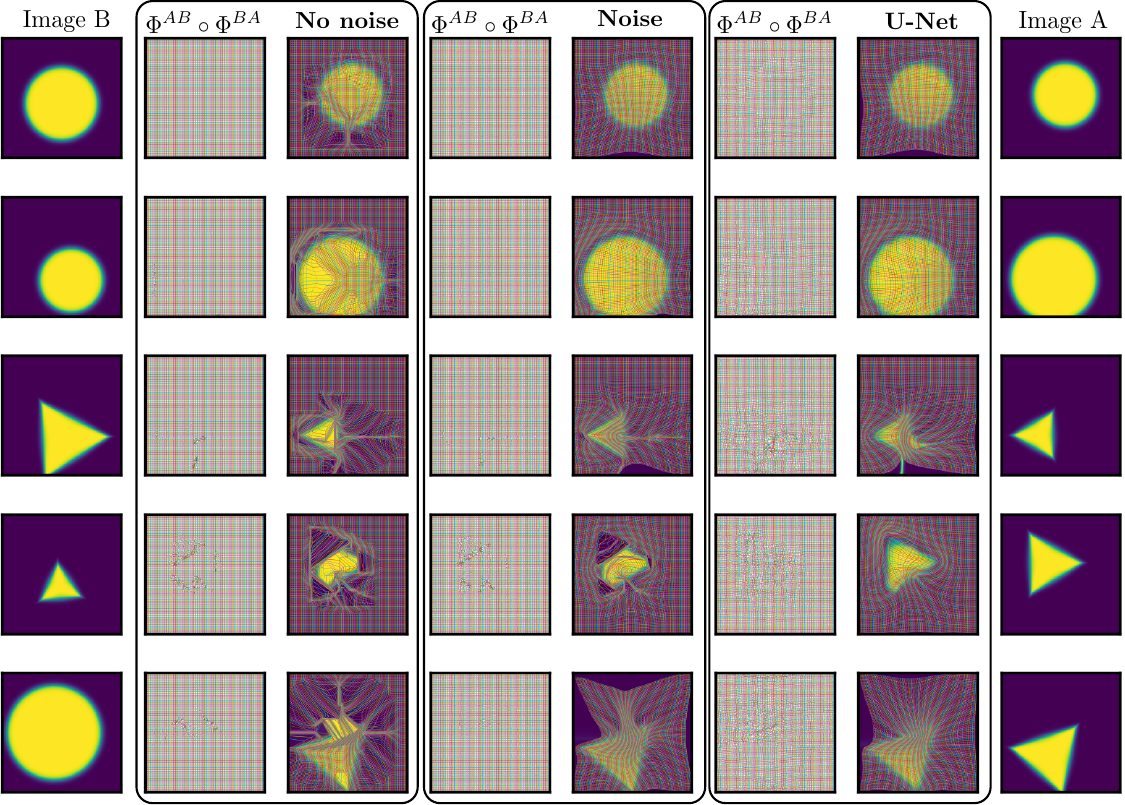}
   }
   \caption{Comparison between \textbf{U-Net} results and \textbf{direct optimization} (no neural network; over $\Phi_\theta^{AB}$ and $\Phi_\theta^{BA}$) w/ and w/o added noise,  using the inverse consistency loss with $\lambda = 2,048$. Direct optimization w/o noise leads to irregular maps, while adding noise or using the \textbf{U-Net} improves map regularity (best viewed zoomed).} \label{fig:regularity_by_inexact_inverse_consistency}
     \vspace{-0.2cm}
\end{figure}

Sec.~\ref{subsec:h1_by_approximate_inverse_consistency} formalized that approximate inverse consistency results in regularizing effects. Specifically, when $\Phi_\theta^{AB}$ is approximately the inverse of $\Phi_\theta^{BA}$, the inverse consistency loss $\mathcal{L}^{AB}_{\text{inv}}$ can be approximated based on Eq.~\eqref{EqH1regularization}, highlighting its implicit $H^1$ regularization.
We investigate this behavior by three experiments: 
\mn{Pair-wise image registration (1) with artificially added noise (\textbf{noise}) and (2) without (\textbf{no noise}) artificially added noise, and (3) population-based registration via a \textbf{U-Net}.}
Fig.~\ref{fig:regularity_by_inexact_inverse_consistency} shows some sample results, supporting our theoretical exposition of Sec.~\ref{subsec:h1_by_approximate_inverse_consistency}: Pair-wise image registration without noise results in highly irregular transformations even though the inverse consistency loss is used. Adding a  small amount of Gaussian noise with standard deviation of 1/8th of a pixel (similar to the inverse consistency loss magnitudes we observe for a deep network) to the displacement fields before computing the inverse consistency loss, results in significantly more regular maps. Lastly, using a \textbf{U-Net} yields highly regular maps. Notably, all three approaches result in approximately inverse consistent maps. The behavior for pair-wise image registration elucidates why inverse consistency has not appeared in the classical (pair-wise) registration literature as a replacement for more complex spatial regularization. The proposed technique \emph{only} results in regularity when inverse consistency errors are present.

\vskip0.5ex
\emph{In summary, our theory is supported by our experimental results: approximate inverse consistency regularizes maps.}



\subsection{Regularization for different networks}

\begin{table}
\centering
\resizebox{\columnwidth}{!}{%
\begin{tabular}{lllllllll}
\toprule
& \multicolumn{8}{c}{\textbf{MNIST}}  \\ 
\midrule
Network $\rightarrow$ & \multicolumn{2}{c}{\bf MLP} & \multicolumn{2}{c}{\bf Enc-Dec} &\multicolumn{2}{c}{\bf U-Net} & \multicolumn{2}{c}{\bf ConvOnly} \\
\midrule
$\lambda$ $\downarrow$ & Dice & Folds~~ & Dice & Folds &  Dice & Folds & Dice & Folds  \\
\midrule
64 & 0.92 & 26.61 & 0.80 & 0.15 & \textbf{0.93} & 3.87 & \textbf{0.93} & 30.20  \\
128 & \textbf{0.92} & 9.95 & 0.77 & 0.08 & \textbf{0.92} & 1.45 & 0.90 & 16.27 \\
256 & \textbf{0.91} & 2.48 & 0.72 & 0.01 & 0.90 & 0.41 & 0.88 & 7.17  \\
512 & \textbf{0.90} & 0.72 & 0.66 & 0.03 & 0.89 & 0.09 & 0.85 & 3.12  \\
1,024 & \textbf{0.88} & 0.34 & 0.62 & 0.06 & 0.86 & 0.02 & 0.81 & 0.54\\
2,048 & \textbf{0.87} & 0.16 & 0.63 & 0.00 & 0.73 & 0.09 & 0.76 & 0.07\\
\bottomrule
\end{tabular}
}\\
\resizebox{\columnwidth}{!}{%
\begin{tabular}{lllllllll}
\toprule
& \multicolumn{8}{c}{\textbf{Triangles \& Circles}} \\ 
\midrule
Network $\rightarrow$ & \multicolumn{2}{c}{\bf MLP } & \multicolumn{2}{c}{\bf  Enc-Dec} &\multicolumn{2}{c}{\bf U-Net} & \multicolumn{2}{c}{\bf ConvOnly} \\
\midrule
$\lambda$ $\downarrow$  & Dice & Folds~~ & Dice & Folds &  Dice & Folds & Dice & Folds  \\
\midrule
64 & \textbf{0.98} & 1.24 & 0.94 & 3.50 & \textbf{0.98} & 2.74 & 0.97 & 12.57  \\
128 & \textbf{0.98} & 0.73 & 0.90 & 2.71 & \textbf{0.98} & 1.59 & 0.96 & 10.15  \\
256 & \textbf{0.98} & 0.27 & 0.88 & 1.11 & 0.97 & 1.14 & 0.96 & 8.49  \\
512 & \textbf{0.97} & 0.10 & 0.87 & 0.65 & 0.96 & 0.70 & 0.94 & 6.61  \\
1,024 & \textbf{0.96} & 0.03 & 0.86 & 0.22 & 0.95 & 0.25 & 0.92 & 3.91  \\
2,048 & \textbf{0.95} & 0.03 & 0.85 & 0.15 & 0.94 & 0.09 & 0.89 & 2.18  \\
\bottomrule
\end{tabular}
}
\caption{Network performance across architectures and regularization strength $\lambda$. \textbf{MLP} / \textbf{U-Net} perform best. All methods work.}\label{tab:registration_across_architectures}
\vspace{-0.2cm}
\end{table}

\begin{figure}
  \includegraphics[width=0.97\columnwidth]{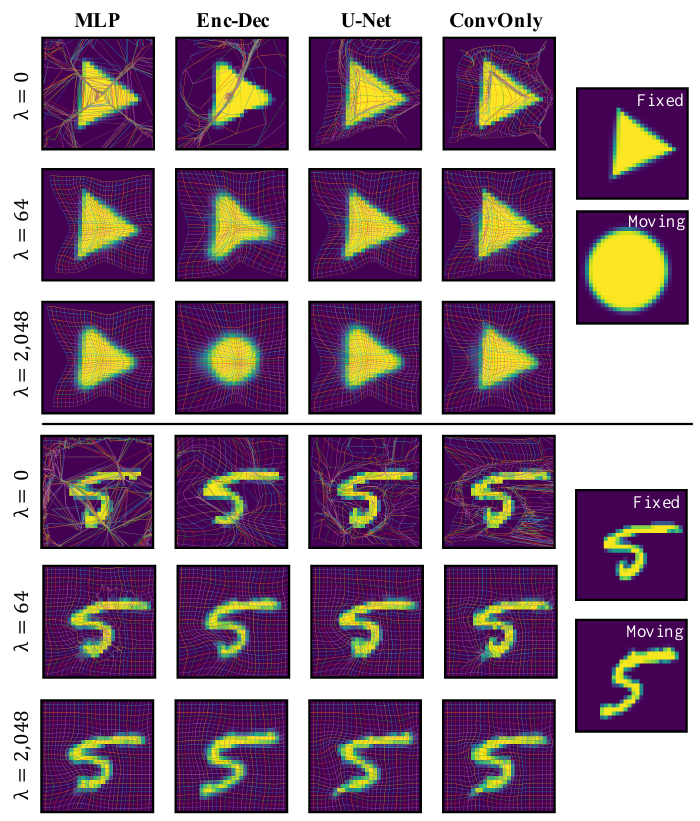}

\caption{Comparison of networks as a function of $\lambda$. \textbf{U-Net} and \textbf{MLP} show the best performance due to their ability to capture long and short range dependencies. \textbf{Enc-Dec} and \textbf{ConvOnly}, which capture only long range and only short range dependencies, resp., also learn regular maps, but for a narrower range of $\lambda$. In all cases, maps become smooth for sufficiently large $\lambda$. Best viewed zoomed.} \label{fig:registration_across_architectures}
\vspace{-0.2cm}
\end{figure}

Sec.~\ref{subsec:imperfect_inverse_consistency} illustrated that approximate inverse consistency yields regularization effects which translate to regularity for network predictions, as networks will, in general, not achieve perfect inverse consistency. A natural next question to ask is ``how much the results depend on a particular architecture''? To this end, we assess four different network types, focusing on MNIST and the triangles \& circles data. We report two measures on held-out images: the \emph{Dice score} of pixels with intensity greater than $0.5$, and the mean number of \emph{folds}, \ie, pixels where the volume form $\mathrm{d}V$ of $\Phi$ is negative.



\vskip0.5ex
One hypothesis as to how network design could drive smoothness 
would be that smoothness is induced by convolutional layers (which can implement a smoothing kernel). 
If this were the case, we would expect the \textbf{MLP} to produce irregular maps with a high number of folds. Vice versa, since the \textbf{MLP} has no spatial prior, obtaining smooth transforms would indicate that smoothness is promoted by the loss itself. The latter is supported by Fig.~\ref{fig:registration_across_architectures}, showing regular maps even for the \textbf{MLP} when $\lambda$ is sufficiently large. Note that $\lambda=0$ in Fig.~\ref{fig:registration_across_architectures} corresponds to an unregularized MSE solution, as discussed in Sec.~\ref{subsection:regularization_thoughts_and_sorting}; maps are, as expected, highly irregular and regularization via inverse consistency is clearly needed.


\vskip0.5ex
A second hypothesis is that regularity results from a \emph{bottleneck} structure within a network, \eg, a \textbf{U-Net}. In fact, Bhalodia \etal~\cite{bhalodia19} show that autoencoders tend to yield smooth maps. To assess this hypothesis, we focus on the \textbf{Enc-Dec} and \textbf{ConvOnly} type networks; the former has a bottleneck structure, while the latter does not.
Fig.~\ref{fig:registration_across_architectures} shows some support for the hypothesis that a bottleneck promotes smooth maps: for a specific $\lambda$, \textbf{Enc-Dec} appears to have more strongly regularized outputs compared to \textbf{U-Net}, with \textbf{ConvOnly} being the most irregular.
Yet, higher values of $\lambda$ (\eg, 1,024 or 2,048) for \textbf{ConvOnly} yield equally smooth maps. Overall, a bottleneck structure does have a regularizing effect, but regularity can also be achieved by appropriately weighing the inverse consistency loss (see Tab.~\ref{tab:registration_across_architectures}).

\vskip0.5ex
\emph{In summary, our experiments indicate that the regularizing effect of inverse consistency is a robust property of the loss, and should generalize well across architectures.}

\subsection{Performance for 3D image registration}


For experiments on real data, we focus on the 3D OAI dataset. To demonstrate the versatility of the advocated inverse consistency loss in promoting map regularity, we refrain from affine pre-registration (as typically done in earlier works) and simply compose \mn{the maps of} \emph{multiple U-Nets} instead. In particular, we compose up to four U-Nets \mn{as follows:} A composition of two U-Nets is initially trained on low-resolution image pairs. Weights are then frozen and this network is composed with a third U-Net, trained on high-resolution image pairs. This network is then optionally frozen and composed with a fourth U-Net, again trained on high-resolution image pairs. During the training of this multi-step approach, the weighting of the inverse consistency loss is gradually increased. We train using ADAM \cite{Kingma15a} with a batch size of 128 in the low-res. stage, and a batch size of 16 in the high-res. stage. MSE is used as image similarity measure.

\vskip0.5ex
We compare our approach, \textcolor{ggreen}{InverseConsistentNet} \mn{(\texttt{ICON})}, against the methods of \cite{shen2019networks}, in terms of (1) cartilage Dice scores between registered image pairs~\cite{AmbellanTackEhlkeetal.2018} (based on manual segmentations) and (2) the number of folds. The segmentations are not used during training and allow quantifying if the network yields semantically meaningful registrations. Tab.~\ref{tab:oai_results} lists the corresponding results, Fig.~\ref{fig:teaser} shows several example registrations. Unlike the other methods in Tab.~\ref{tab:oai_results}, except where explicitly noted, \mn{\texttt{ICON} does not require affine pre-registration. Since affine maps are inverse consistent, they are not penalized by our method.}
Notably, despite its simplicity, \texttt{ICON} yields performance (in terms of Dice score \& folds) comparable to more complex, explicitly regularized methods. We emphasize that our objective is not to outperform existing techniques, but to present evidence that regular maps can be learned \emph{without} carefully tuned regularizers.

\vskip0.5ex
\emph{In summary, using the proposed inverse consistency loss yields (1) competitive Dice scores, (2) acceptable folds, and (3) fast performance.}



\begin{table}
\begin{small}
\centering
\resizebox{\columnwidth}{!}{%
\begin{tabular}{lllll}
\toprule
\textbf{Method} &  $\mathcal{L}_{sim}$ & \textbf{Dice} & \textbf{Folds} & \textbf{Time} [s]\\
\midrule
Demons & MSE & 63.47 & 19.0 & 114\\
SyN & CC & 65.71 & 0 & 1330\\
NiftyReg & NMI & 59.65 & 0 & 143\\
NiftyReg & LNCC & 67.92 & 203 & 270\\
vSVF-opt & LNCC & 67.35 & 0 & 79\\
Voxelmorph (w/o affine) & MSE & 46.06 & 83 & 0.12\\
Voxelmorph & MSE & 66.08 & 39.0 & 0.31\\
AVSM (7-Step Affine, 3-Step Deformable) & LNCC & 68.40 & 14.3 & 0.83\\
\midrule
\textcolor{ggreen}{\texttt{ICON}} (2 step \nicefrac{1}{2} res., 2 step full res., w/o affine) & MSE & 68.29 & 118.4 & 1.06\\
\textcolor{ggreen}{\texttt{ICON}} (2 step \nicefrac{1}{2} res., 1 step full res., w/o affine) & MSE & 66.16 & 169.4  & 0.57 \\
\textcolor{ggreen}{\texttt{ICON}} (2 step \nicefrac{1}{2} res., w/o affine) & MSE & 59.36 & 49.35  & 0.09 \\
\bottomrule 
\end{tabular}
}
\caption{Comparison of \textcolor{ggreen}{\texttt{ICON}} against the methods in \cite{shen2019networks}, on cross-subject registration for OAI knee images.\label{tab:oai_results}}
\vspace{-0.2cm}
\end{small}
\end{table}

\section{Limitations, future work, \& open questions}
\label{sec:future_work}

Several questions remain and there is no shortage of theoretical/practical directions, some of which are listed next.


\vskip0.5ex
\noindent
\textbf{Network architecture \& optimization.} Instead of specifying a spatial regularizer, we now specify a network architecture. While our results suggest regularizing effects for a variety of architectures, we are still lacking a clear understanding of how network architecture and numerical optimization influence solution regularity.

\vskip0.5ex
\noindent
{\bf Diffemorphisms at test time.} We simply encourage inverse consistency via a quadratic penalty. Advanced numerical approaches (\eg, augmented Lagrangian methods~\cite{nocedal2006numerical}) could more strictly enforce inverse consistency during \emph{training}. Our current approach is only \emph{approximately diffeomorphic} at test time. To guarantee diffeomorphisms, one could explore combining inverse consistency with fluid deformation models~\cite{holden2007review}. These have been used for deep registration networks~\cite{yang2017quicksilver,yang2016fast,shen2019networks,shen2019region,dalca2018unsupervised} combined with explicit spatial regularization. We would simply predict a velocity field and obtain the map via integration. By using our loss, sufficiently smooth velocity fields would likely emerge. Alternatively, one could use diffeomorphic transformation parameterizations by enforcing positive Jacobian determinants~\cite{shu2018deforming}.

\vskip0.5ex
\noindent
{\bf Multi-step.} Our results show that using a multi-step estimation approach is beneficial; successive networks can refine deformation estimates and thereby improve registration performance. What the limits of such a multi-step approach are (i.e., when performance starts to saturate) and how it interacts with deformation estimates at different resolution levels would be interesting to explore further. 

\vskip0.5ex
\noindent
{\bf Similarity measures.} For simplicity, we only explored MSE. NCC, local NCC, and mutual information would be natural choices for multi-modal registration. In fact, there are many opportunities to improve registrations \eg using more discriminative similarity measures based on network-based features, multi-scale information, or side-information during training, \eg, segmentations or point correspondences.

%

\vskip0.5ex
\noindent
{\bf Theoretical investigations.} It would be interesting to establish how regularization by inverse consistency relates to network capacity, expressiveness, and generalization. Further, establishing a rigorous theoretical understanding of the regularization effect due to the data \emph{population} and its link with inverse consistency would be important.

\vskip0.5ex
\noindent
{\bf General inverse consistency.} Our work focused on spatial correspondences for registration, but the benefits of inverse consistency regularization are likely much broader. For instance, its applicability to general mapping problems (\eg, between feature vectors) should be explored.

\section{Conclusion}
\label{sec:conlusion}

We presented a deliberately simple deep registration model which generates approximately diffeomorphic maps by regularizing via an inverse consistency loss. We theoretically analyzed why inverse consistency leads to spatial smoothness and empirically showed the effectiveness of our approach, yielding \mn{competitive} 
3D registration performance. 

\vskip1ex
Our results suggest that simple deep registration networks might be as effective as more complex approaches which require substantial hyperparameter tuning and involve choosing complex transformation models. As a wide range of inverse consistency loss penalties lead to good results, only the desired similarity measure needs to be chosen and extensive hyperparameter tuning can be avoided. This opens up the possibility to easily train extremely fast custom registration networks on given data. Due to its simplicity, ease of use, and computational speed, we expect our approach to have significant practical impact. We also expect that inverse consistency regularization will be useful for other tasks, which should be explored in future work.


\opt{for_gig,icml_accepted,iccv_submission}
{
  \section*{Acknowledgments}

  This research was supported in part by Award Number R21-CA223304 from the National Cancer Institute and Award Number 1R01-AR072013 from the National Institute of Arthritis and Musculoskeletal and Skin Diseases of the National Institutes of Health. It was also supported by an MSK Cancer Center Support Grant/Core Grant P30 CA008748 and by the National Science Foundation (NSF) under award number NSF EECS-1711776. It was also supported by the Austrian Science Fund (FWF): project FWF P31799-N38 and the Land Salzburg (WISS 2025) under project numbers 20102- F1901166-KZP and 20204-WISS/225/197-2019. The content is solely the responsibility of the authors and does not necessarily represent the official views of the NIH or the NSF. The authors have no conflicts of interest.
}
  
\opt{for_gig,icml_submission}
{
  \bibliographystyle{icml2021}
  \bibliography{mybibliography}
}
\opt{iccv_submission}
    {
      
      {\small
        \bibliographystyle{ieee_fullname}
        \bibliography{mybibliography}
      }
    }

\end{document}


\sloppy

\opt{iccv_submission}{
\title{ICON: Supplementary Material}

\author{Hastings Greer\\
  Department of Computer Science\\
UNC Chapel Hill, USA\\
{\tt\small tgreer@cs.unc.edu}
\and
Roland Kwitt\\
Department of Computer Science\\
University of Salzburg, Austria\\
{\tt\small Roland.Kwitt@sbg.ac.at}
\and
Fran\c{c}ois-Xavier Vialard\\
LIGM, Universit\'e Gustave Eiffel, France\\
{\tt\small francois-xavier.vialard@u-pem.fr}
\and
Marc Niethammer\\
 Department of Computer Science\\
UNC Chapel Hill, USA\\
{\tt\small mn@cs.unc.edu}
}

\maketitle
\ificcvfinal\thispagestyle{empty}\fi
}
\thispagestyle{plain}
\pagestyle{plain}

\opt{for_gig,icml_submission}{

\icmltitlerunning{Learning Regular Maps Through Inverse Consistency}
  
\onecolumn[
\icmltitle{Learning Regular Maps Through Inverse Consistency}

\icmlsetsymbol{equal}{*}

\begin{icmlauthorlist}
\icmlauthor{Hastings Greer}{unc}
\icmlauthor{Roland Kwitt}{salzburg}
\icmlauthor{Fran\c{c}ois Xavier-Vialard}{upem}
\icmlauthor{Marc Niethammer}{unc}
\end{icmlauthorlist}

\icmlaffiliation{unc}{Department of Computer Science, University of North Carolina at Chapel Hill, USA}
\icmlaffiliation{salzburg}{Department of Computer Science, University of Salzburg, Austria}
\icmlaffiliation{upem}{LIGM, Universit\'e Gustave Eiffel, France}

\icmlcorrespondingauthor{Hastings Greer}{tgreer@cs.unc.edu}

\icmlkeywords{Deep Learning, Image Registration, Inverse Consistency}

\vskip 0.3in
]

\printAffiliationsAndNotice{\icmlEqualContribution} 
}

\sloppy

\section*{Introduction}
We cover these topics in the supplementary material that did not fit into the main manuscript:
\begin{itemize}
\item We retrace the proof of the $H^1$ regularizing property of approximate inverse consistency in greater detail.
\item We specify all details of the neural network architectures used in the manuscript's experiments, including number of features, number of layers, training proceedure, etc.
\end{itemize}

\section{Detailed proof of the regularizing effect of inverse consistency}

This section details our derivation for the smoothness properties emerging from approximate inverse consistency. 


Denote by $\fxvarphi(x)$ the output map of a network for images $(I^A,I^B)$ and by $\fxpsi(x)$ the output map between $(I^B,I^A)$
Recall that we add two independent spatial white noises $n_1(x),~n_2(x)\in\mathbb{R}^N$ ($x \in [0,1]^N$ with $N = 2$ or $N = 3$ the dimension of the image)
of variance $1$ for each spatial location to the two output maps and define $\fxvarphivarepsilon(x) := \fxvarphi(x) + \varepsilon n_1(\fxvarphi(x))$ and $\fxpsivarepsilon(x) := \fxpsi(x) + \varepsilon n_2(\fxpsi(x))$ with $\varepsilon$ a positive parameter.
We consider the following loss
\begin{multline}\label{EqLossSymmetric2}
    \mathcal{L} =\lambda \left( \| \fxvarphivarepsilon \circ \fxpsivarepsilon - \operatorname{Id} \|^2_2 + \| \fxpsivarepsilon \circ \fxvarphivarepsilon - \operatorname{Id} \|^2_2 \right) + \| I^A \circ \fxvarphi - I^B \|^2_2 + \| I^B \circ \fxpsi  - I^A \|^2_2\,.
\end{multline}
Throughout this section, we give the details of the expansion in $\varepsilon$ of the loss, thus we use the standard notations $o$ and $O$ w.r.t $\varepsilon \rightarrow 0$.
We focus on the first two terms (that we denote by $\lambda \mathcal{L}_{\text{inv}}$) since the regularizing property comes from the inverse consistency.
We expand one of the first two terms  of \eqref{EqLossSymmetric2} since by symmetry the other is similar. If the noise is bounded (or with high probability in the case of Gaussian noise), we have
\begin{equation}
  \| \fxvarphivarepsilon \circ \fxpsivarepsilon - \operatorname{Id} \|^2_2 =   \| \fxvarphi \circ \fxpsi + \varepsilon n_1(\fxvarphi \circ \fxpsi)  + d\fxvarphivarepsilon(\varepsilon n_2(\fxpsi)) - \operatorname{Id}\|^2_2 + o(\varepsilon^2)\,,
\end{equation}
where $d\Phi$ denotes the Jacobian of $\Phi$. By developing the squares and taking expectation, we get 
\begin{multline}\label{EqExpectation}
    \mathbb{E}[\| \fxvarphivarepsilon \circ \fxpsivarepsilon - \operatorname{Id} \|^2_2] = \| \fxvarphi \circ \fxpsi - \operatorname{Id} \|^2_2 + \varepsilon^2 \mathbb{E}[\|n_1\circ(\fxvarphivarepsilon \circ \fxpsivarepsilon)\|^2_2] + \varepsilon^2 \mathbb{E}[\|d\fxvarphivarepsilon( n_2) \circ \fxpsi \|^2_2] +  o(\varepsilon^2)\,,
\end{multline}
since by independence all the cross-terms vanish. Indeed, the noise terms have $0$ mean value.
The second term is constant:
\begin{multline}
    \mathbb{E}[\|n_1\circ(\fxvarphivarepsilon \circ \fxpsivarepsilon)\|^2_2] =
     \mathbb{E}[\int \| n_1\|^2_2(y)\operatorname{Jac}((\fxpsivarepsilon)^{-1} \circ (\fxvarphivarepsilon)^{-1}) dy]  \\
     = \int \mathbb{E}[\|n_1\|^2_2(y)] \operatorname{Jac}((\fxpsivarepsilon)^{-1} \circ (\fxvarphivarepsilon)^{-1})~dy= const \,, \nonumber
\end{multline}
where we performed a change of variables and denoted the determinant of the Jacobian matrix as $\operatorname{Jac}$. The last equality follows from the fact that $\mathbb{E}[\|n_1\|^2_2(y)] = 1 \,\forall y$, \ie the variance of the noise is constant equal to $1$. Last, we also use the change of variables $y = \fxvarphivarepsilon \circ \fxpsivarepsilon(x)$.
By similar computations, the last term in Equation \eqref{EqExpectation} is equal to
\begin{equation}\label{EqWhiteNoise}
  \mathbb{E}[\|d\fxvarphivarepsilon( n_2) \circ \fxpsi \|^2_2] =   \int \mathbb{E}[(n_2^{\top}d(\fxvarphivarepsilon)^{\top} d\fxvarphivarepsilon(n_2)) \circ \fxpsi]\,dx\,.
  \end{equation}
  In the next formula, we use coordinate notations. For $i,k \in 1,\ldots , N$, we denote by $\partial_i \Psi^k$ the partial derivative w.r.t. the i$^{\text{th}}$ coordinate of the k$^{\text{th}}$ component of the map $\Psi: \mathbb{R}^N \to \mathbb{R}^N$, the notation $n^i(x)$ stands for the i$^{\text{th}}$ component of the noise, $i \in 1,\ldots , N$. 
Using these notations, we have 
  \begin{multline}\label{EqTraceSimplification}
     \mathbb{E}[(n_2(x))^{\top}d(\fxvarphivarepsilon)^{\top} d\fxvarphivarepsilon(n_2(x))]=\mathbb{E}[ \sum_{k,i,j} n_2^i(x)\partial_i [\fxvarphivarepsilon]^k(x) \partial_j [\fxvarphivarepsilon(x)]^k n_2^j(x)]
     \\
     =\mathbb{E}[ \sum_{k,i} \partial_i [\fxvarphivarepsilon]^k(x) \partial_i [\fxvarphivarepsilon(x)]^k]
     \,.
  \end{multline}
   In the previous equation, we used the property of the white noise $n_2$ which satisfies null correlation in space and dimension $\mathbb{E}[n_2^k(x)n_2^{k'}(x')] = 1$ if $(k,x)=(k',x')$ and $0$ otherwise. Recognizing the trace in Formula \eqref{EqTraceSimplification}, we finally get
  \begin{equation}\label{EqWhiteNoise2}
\mathbb{E}[\|d\fxvarphivarepsilon( n_2) \circ \fxpsi \|^2_2]  = \int \operatorname{Tr}([d(\fxvarphivarepsilon)^{\top} d\fxvarphivarepsilon] \circ \fxpsi) dx 
   = \int \operatorname{Tr}(d(\fxvarphivarepsilon)^{\top} d\fxvarphivarepsilon) \operatorname{Jac}((\fxpsi)^{-1})~dy\,,
\end{equation}
where $\operatorname{Tr}$ is the trace. 
The last equality follows from the change of variables with $\fxpsi$.

\par{\textbf{Approximation and resulting $H^1$ regularization: }}
Under the hypothesis that $\fxvarphi \circ \fxpsi,\fxpsi \circ \fxvarphi$ are close to identity, one has 
$\operatorname{Jac}((\fxpsi)^{-1}) = \operatorname{Jac}(\fxvarphi) + o(1)$. Therefore, the last term of \eqref{EqWhiteNoise2} is approximated by
\begin{equation}\label{EqFirstSimplification}
   \int \operatorname{Tr}(d(\fxvarphivarepsilon)^{\top} d\fxvarphivarepsilon) \operatorname{Jac}((\fxpsi)^{-1})~dy = 
   \int \operatorname{Tr}(d(\fxvarphivarepsilon)^{\top} d\fxvarphivarepsilon) \operatorname{Jac}(\fxvarphi)~dy + o(1)\,.
\end{equation}
Note that we only need an approximation at zero$^{\text{th}}$ order, since this term appears at second order in the penalty $\mathcal{L}_{\text{inv}}$.
Assuming this approximation holds
, we use it in Eq.~\eqref{EqWhiteNoise2}, together with the fact that, $\fxvarphivarepsilon = \fxvarphi + O(\varepsilon)$ to approximate at order $\varepsilon^2$ the quantity $\mathcal{L}_{\text{inv}}$, \ie,
\begin{equation}
  \mathcal{L}_{\text{inv}}  = \left\| \fxvarphi \circ \fxpsi - \operatorname{Id}\right\|^2_2  + \left\| \fxpsi \circ \fxvarphi - \operatorname{Id}\right\|^2_2 
   + \varepsilon^2 \left\|  d\fxvarphi \sqrt{\operatorname{Jac}(\fxvarphi)} \right\|^2_2 
   + \varepsilon^2 \left\|  d\fxpsi \sqrt{\operatorname{Jac}(\fxpsi)} \right\|^2_2 + o(\varepsilon^2) \, .
\label{EqH1regularization}
\end{equation}
Last, the square root that appears in the $L^2$ norm is simply a rewriting of the term on the r.h.s. of Eq.~\eqref{EqFirstSimplification}. 


We see that approximate inverse consistency leads to an $L^2$ penalty of the gradient, weighted by the Jacobian of the map. Generally, this is a type of Sobolev ($H^1$ more precisely) regularization sometimes used in image registration in a different context, see \cite{template_matching} for a particular instance. In particular, the $H^1$ term is likely to control the compression and expansion magnitude of the maps, at least on average, on the domain.
Hence, approximate inverse consistency leads to an implicit $H^1$ regularization, formulated directly on the map. 
\par
In comparison with the literature, the regularization is not formulated on the velocity field defining the displacement by integration as it is standard in diffeomorphic registration of pairwise images since the pioneering work of \cite{DuGrMi1998}. In our context, the resulting $H^1$ penalty concerns the map itself. On the theoretical side, one can ask if such regularization makes the problem well posed from the analytical point of view, \ie existence of minimizers, regularity of solutions. However, few works have explored this type of regularization directly on the map, see for instance the work in \cite{DEPASCALE2016237} in the context of optimal transport. In contrast, $H^1$ regularization of the velocity field has been explored resulting in a non-degenerate metric on the group of diffeomorphisms as proven in \cite{Michor2005}.
\par{\textbf{Further discussion: }}
When the noise level $\varepsilon$ is turned to $0$, we also observe a regularizing effect when the map is output by a neural network. (Although not when the map is directly optimized.) Since the network does not perfectly satisfy the inverse consistency soft constraint, we conjecture that the resulting error behaves like the white-noise we studied above, thereby explaining the observed regularization. 
\par
Another important challenge is to understand the regularization bias which comes from the population effect. In this case, we conjecture that this approach makes learning the regularization metric
more adaptive to the given population data.
\vskip0.5ex
\emph{However, a fully rigorous theoretical understanding of the regularization effect due to the data population and its link with inverse consistency when no noise is used 
is important, but beyond our scope here.}

\section{Network Architectures}

In this manuscript we refer to four neural network architectures: MLP, Encoder-Decoder, U-Net, and Convolutions. The details of each are provided next.\\
\subsection{MLP} MLP refers to a multilayer perceptron with no special structure. The input, a pair of images, is flattened into a vector. Next, it is passed through hidden layers of size 8000 and 3000, each followed by a ReLU nonlinearity. The output layer is of dimension 2 $\cdot$ Image Width $\cdot$ Image Height, which is reshaped into a grid of displacement vectors from which $\Phi$ is calculated.\\
\subsection{Encoder-Decoder} The Encoder-Decoder used in this manuscript is composed of a convolutional encoder and decoder, resembling a U-Net with no skip connections. Each layer consists of a stride 2 convolution or transpose convolution with kernel size 3x3 in the encoder and 4x4 in the decoder, followed by a ReLU nonlinearity. The layers have 16, 32, 64, 256, and 512 features in the encoder, and 256, 128, 64, 32, and 16  features in the decoder. As in all cases, the output is a grid of displacement vectors.\\
\subsection{ConvOnly} This refers to an architecture consisting of six convolutional layers, each with 5x5 kernel, ReLU nonlinearity,  and 10 output features. No downsampling or upsampling is performed. Each layer is fed as input the concatenation of the outputs of all previous layers. \\ 
\subsection{U-Net} This is a U-Net with skip and residual connections. The convolutional layers have the same shapes and output dimensions as the encoder decoder network, but use Leaky ReLU activation placed before convolution instead of after. In addition, batch normalization is inserted before each convolution, and a residual connection is routed around each convolution, using upsampling or downsampling as required to match the image size. \par

For all four of these architectures the weights of the final layer of the neural network are initialized to zero instead of randomly, such that training begins with the network outputting a displacement field of zero. The code specifying these architectures is included in the file \path{networks.py}

\section{Software Architecture}
 In the codebase that we developed for this paper, registration algorithms are implemented as subclasses of \verb|pytorch.nn.Module|. A registration algorithm's \verb|forward| method takes as input a batch of pairs of images to register $I^A$ and $I^B$, and returns a python function $\Phi[I^A, I^B]:\mathbb{R}^N \rightarrow \mathbb{R}^N$ that maps a batch of vectors from the space of image B to the space of image A. For example, to use a neural network that outputs a displacement field as a registration algorithm, we wrap it in the class FunctionFromVectorField:

\begin{lstlisting}[language=Python]
class FunctionFromVectorField(nn.Module):
    def __init__(self, net):
        super(FunctionFromVectorField, self).__init__()
        self.net = net

    def forward(self, image_A, image_B):
        vectorfield_phi = self.net(image_A, image_B)

        def ret(input_):
            return input_ + compute_warped_image_multiNC(
                vectorfield_phi, input_, self.spacing, 1
            )
        return ret
\end{lstlisting}
where \path{compute_warped_image_multiNC} interpolates between the vectors of its first argument at tensor of positions specified by its second argument.
This code corresponds to equation (11) in the manuscript. Note especially that in the returned function \verb|ret|, we add the input to the interpolated displacement. We do not attempt to interpolate a grid representation of a map (ie, a voxelized displacement field added to a voxelized identity map), as a displacement field can be extrapolated naturally, but a map cannot.   

We find this organizational convention to be highly composable: this approach makes it simple to construct a registration algorithm that expands on the behavior of a component registration algorithm. For example, to operate on a high resolution pair of images with a low resolution registration algorithm, we use a wrapper with the following simple \verb|forward| method:

\begin{lstlisting}[language=Python]
def forward(self, image_A, image_B):
    x = self.avg_pool(x, 2, ceil_mode=True)
    y = self.avg_pool(y, 2, ceil_mode=True)
    return self.wrapped_algorithm(x, y)
\end{lstlisting}
Since the output is a fully fledged function $:\mathbb{R}^N \rightarrow \mathbb{R}^N$ which is resolution agnostic, it does not need to be modified by this method and can simply be passed along.

\section{Composition of transforms}
We have defined a registration algorithm as a functional $\Phi$ which takes as input two functions $\mathbb{R}^N \rightarrow \mathbb{R}$,
$I^{A}$ and $I^B$, and outputs a map $\mathbb{R}^N \rightarrow \mathbb{R}^N$ that aligns them, specifically satisfying $I^A \circ \Phi[I^A, I^B] \simeq I^B$.
Most registration algorithms have multiple steps, such as an affine step followed by a deformable step, and so it is useful to define how to compose two algorithms (i.e., two procedures for computing a map from a pair of images) $\Phi$ and $\Psi$. 
The most obvious approach to this problem is to apply $\Phi$ to the problem $\{I^A, I^B\}$, yielding a function  $\Phi^{AB}$ such that $I^A \circ \Phi^{AB} \simeq I^B$. Then, the intermediate image $\tilde{I}^A := I^A \circ \Phi^{AB}$ is computed using the function $\Phi^{AB}$ that was found, and the second registration problem is declared to be registering $\{\tilde{I}^A, I^B\}$ by computing a map using the algorithm $\Psi$.
Putting this together, we set out to define an operator TwoStep satisfying the equation
\begin{equation}
I^A \circ TwoStep\{\Phi, \Psi\}[I^{A}, I^B] = (\tilde{I}^{A}) \circ \Psi[\tilde{I}^A, I^B] \simeq I^B \,,
\end{equation}
\begin{equation}
I^A \circ TwoStep\{\Phi, \Psi\}[I^{A}, I^B] = (I^{A} \circ \Phi[I^A, I^B]) \circ \Psi[I^A \circ \Phi[I^A, I^B], I^B] \simeq I^B \,.
\end{equation}
Since composition is associative, we can move the parentheses and isolate TwoStep as 
\begin{equation}
TwoStep\{\Phi, \Psi\}[I^{A}, I^B] = \Phi[I^A, I^B] \circ \Psi[I^A \circ \Phi[I^A, I^B], I^B] \,.
\end{equation}
    
\noindent This is implemented in our code base as another registration algorithm with the following forward method 
\begin{lstlisting}[language=Python]
def forward(self, image_A, image_B):
    phi = self.netPhi(image_A, image_B)
    phi_vectorfield = phi(self.identityMap)
    self.image_A_comp_phi = compute_warped_image_multiNC(
        image_A, phi_vectorfield, self.spacing, 1
    )
    psi = self.netPsi(self.image_A_comp_phi, image_B)
    return lambda input_: phi(psi(input_))
\end{lstlisting}

\section{Training Procedure for synthetic data and MNIST}
\subsection{Regularization by approximate inverse consistency}
To investigate the regularizing effects of approximate inverse consistency, we register an image of a circle and a triangle by three methods: By directly optimizing the forward and reverse displacement vector fields, by directly optimizing the displacement vector fields with added noise, and by optimizing a U-Net that outputs a displacement vector field over a dataset that includes the image of a circle and triangle. This dataset was generated as follows: 
a center $(cx, cy)$ for each image is sampled uniformly from $[0.4, 0.7] \times [0.4, 0.7]$, a radius $r$ is sampled from $[0.2, 0.4]$, and an angle $\theta$ from $[0, 2 \pi]$. Each point in the image is then associated with an intensity by one of the following formulas:
Half of the generated images are chosen to be circles and their intensities are set via the expression
\begin{equation}
    \tanh\left(- 40 \cdot (\sqrt{(x - cx)^2 + (y - cy)^2} - r)\right) \,,
\end{equation}
while the remainder are chosen to be triangles and their intensities are set to
\begin{equation}
\tanh\left(-40 \cdot (\sqrt{(x - cx)^2 + (y - cy)^2}
    - r \cdot \frac{\cos(\frac{\pi}{3}) }{ \cos((\arctan2(x - cx, y - cy) + \theta) \% (\frac{2 \pi}{ 3}) - \frac{\pi}{ 3})}\right) \,.
\end{equation}
In each case, the ADAM optimization algorithm is chosen, with a learning rate of $0.0001$. While training the U-Net, a batch size of 128 is used. The code to train the U-Net is included as \path{training_scripts/triangle_example.py}, and the code to directly optimize the deformation fields is included as \path{notebooks/NoNetwork.ipynb}
\subsection{Regularization for different networks}
For all architectures and choices of $\lambda$, the following training procedure was followed. The network is trained in pytorch using the Adam algorithm, a batch size of 128, and a learning rate of $0.0001$, for 18750 steps (400 epochs). The self contained notebook to generate/download the datasets, train each combination and generate Figure 6 is included in the supplementary files as \path{notebooks/InverseConsistencyGenerateFigure6.ipynb}. This code can be downloaded and run on its own, as it only depends on pytorch and matplotlib.
\section{Training Procedure for our OAI knee results}

\subsection{Automatic increase of $\lambda$ during training}
During our initial experiments with training our registration network on real data, we found that, in the event that an initial value of $\lambda$ was selected that was too low, leading to an unacceptable degree of folding, we were able to increase $\lambda$ and continue training the network, suppressing the folds without significantly reducing the achieved DICE score. However, when we repeated the training with $\lambda$ beginning at this high value, training proceeded much more slowly due to the ill-conditioned nature of solving a constrained optimization problem using a large quadratic penalty. This was never an issue when registering 2-D datasets, because it was feasible to train on them for a sufficient number of iterations for Adam optimization to automatically resolve the ill conditioning using a small step size. However, on the 3-D dataset this issue threatened to make training times impractical.  Our initial solution to this problem was to begin training with $\lambda$ small, and manually increase $\lambda$ during training whenever the number of folds became large. While this worked well, it introduced a large number of hyperparameters in the form of a complex training schedule, defeating the purpose of our approach. Instead, we decided to select as a hyperparameter the acceptable number of folds, and increase $\lambda$ at each iteration of training if the number of folds measured that iteration exceeded the decided-upon acceptable number of folds. For our low resolution training, $\lambda$ was increased by $.1$ whenever the acceptable folds threshold was exceeded. For our high resolution training, $\lambda$ was increased by $.8$ whenever the acceptable folds threshold was exceeded.
\begin{figure}
    \centering
    \includegraphics{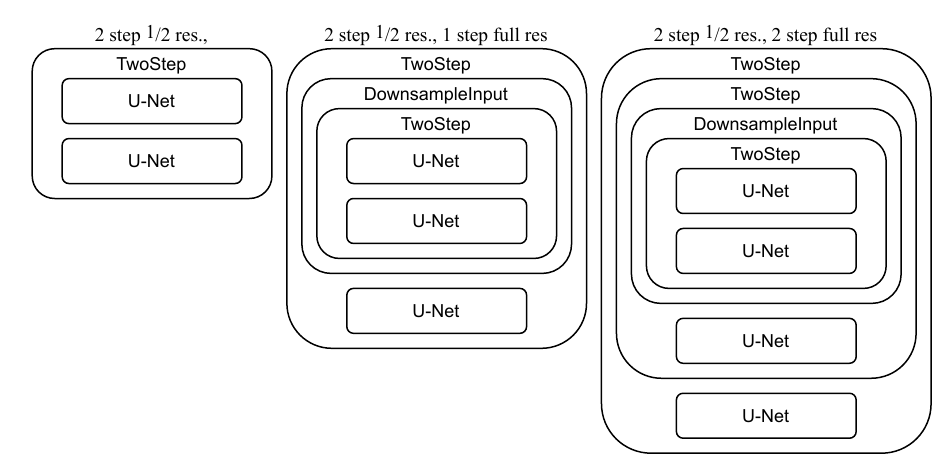}\\
    \caption{Architectures used for our OAI results. The low resolution component inside DownsampleInput only requires an 8th the memory and computing power of the whole network, and pretraining it alone makes our overall approach computationally feasible on a single 4 GPU machine. Once it is pretrained, it is plugged into the larger model as shown.}
    \label{fig:oai-arc}
\end{figure}

\subsection{Details}

The ``acceptable number of folds" hyperparameter was set to 200. 200 was the first value tried for this hyperparameter, however this choice was informed by the outcome of previous experiments where $\lambda$ was set manually. 

First, the `low resolution network' is composed of two U-Nets that each take as input a pair of knee images at half resolution, and output a displacement map. These are combined using the operator TwoStep as described above. The low resolution network is trained end to end with $\lambda$ incremented whenever the batch-mean number of folds exceeds 200, as described above. The batch size used is 128, the learning rate is set to $0.00005$, and the network is trained for 16,000 steps. This low resolution pretraining serves to greatly reduce the overall time needed to train the neural network, since much larger batches of images can fit into GPU memory. This step is performed by the included script \path{training_scripts/double_deformable_knee.py}, and the resulting loss curve is reproduced here in Figure 2.

Second, the `low resolution network' is wrapped with a class that downsamples input images, and then combined with a U-Net that takes as input full resolution images, again using the operator TwoStep. The weights of the low resolution network are frozen, and the full resolution network is trained for 75,000 steps, with a learning rate of $0.00005$ and a batch size of 16. This step is performed by the included script \path{training_scripts/hires_finetune_frozen_lowres.py}. The loss curve associated with this step is presented in Figure 2.
Finally, evaluation of the low resolution and full networks is performed using the included notebooks.  \path{DoubleDeformableDICE.ipynb} and \path{DoubleDeformable-HiresDICE.ipynb} respectively. Training was done on a machine with 4 RTX 3090 GPUs, taking 2 days for the low resolution component and 4 days for the high resolution component.
\begin{figure}
    \centering
    Low Resolution Pretraining\par\medskip
    \includegraphics[width=150pt]{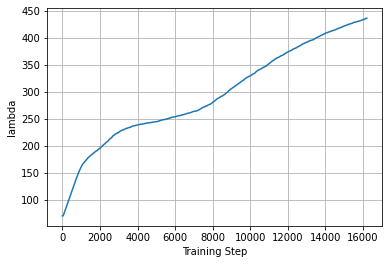}
    \includegraphics[width=150pt]{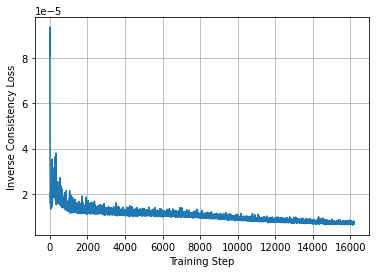}
    \includegraphics[width=150pt]{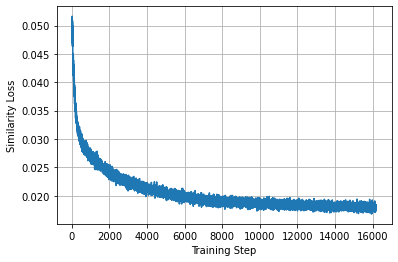} \\
    High Resolution Fine Tuning\par\medskip
    \includegraphics[width=150pt]{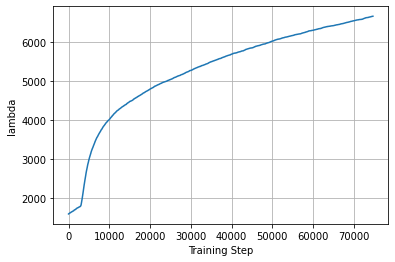}
    \includegraphics[width=150pt]{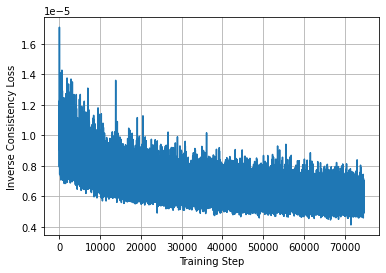}
    \includegraphics[width=150pt]{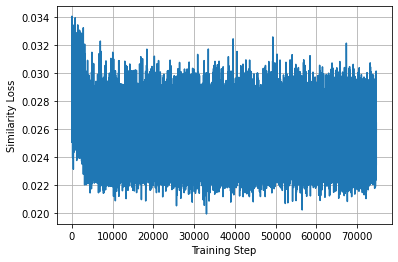} \\
    High Resolution Fine Tuning (Second Step)\par\medskip
    \includegraphics[width=150pt]{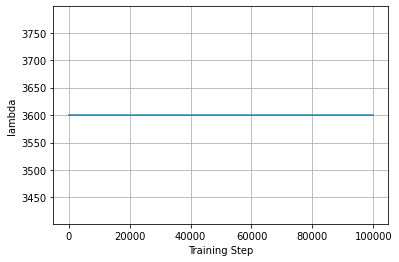}
    \includegraphics[width=150pt]{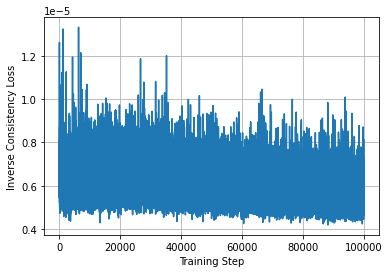}
    \includegraphics[width=150pt]{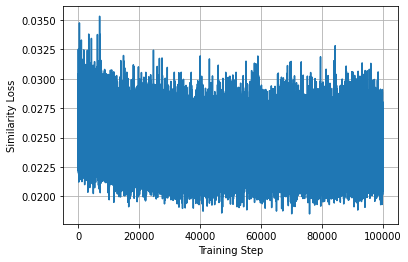}
    \caption{Training curves for our result on OAI dataset. It is interesting that the required value of $\lambda$ to suppress folding increases over the course of training, and in particular increases rapidly once we begin training in high resolution. Nonetheless, our approach of incrementing $\lambda$ by a fixed amount whenever the number of folds in a batch exceeds a threshold successfully generates smooth transforms.}
    \label{fig:training_curves}
\end{figure}

\opt{for_gig,icml_submission}
{
  \bibliographystyle{icml2021}
  \bibliography{mybibliography}
}
\opt{iccv_submission}
    {
      
      {\small
        \bibliographystyle{ieee_fullname}
        \bibliography{mybibliography}
      }
    }